\documentclass[11pt]{article}
\usepackage{fullpage}
\usepackage{cite}

\usepackage{latexsym}
\usepackage{amssymb}
\usepackage{amsmath}
\usepackage{graphicx}

\newcommand{\argmax}{\mbox{argmax}}
\newcommand{\citep}{\cite}

\def\ChapterPath{.}

\newcommand{\subfour}[1]{\vspace*{3mm}\hspace{-3.5mm}{\bf #1}.}

\newcommand{\commentout}[1]{}
\newcommand{\mycomment}[1]{{\bf[[#1]]}}

\newcommand{\TBox}{\mathit{Box}}
\newcommand{\Truck}{\mathit{Truck}}
\newcommand{\City}{\mathit{City}}

\newcommand{\london}{\mathit{london}}
\newcommand{\paris}{\mathit{paris}}
\newcommand{\truck}{\mathit{truck}}
\newcommand{\tbox}{\mathit{box}}
\newcommand{\load}{\mathit{load}}

\newcommand{\unload}{\mathit{unload}}
\newcommand{\unloadS}{\mathit{unloadS}}
\newcommand{\unloadF}{\mathit{unloadF}}
\newcommand{\drive}{\mathit{drive}}

\newcommand{\noop}{\mathit{noop}}

\newcommand{\BIn}{\mathit{BIn}}
\newcommand{\TIn}{\mathit{TIn}}
\newcommand{\On}{\mathit{On}}
\newcommand{\Regr}{\mathit{Regr}}

\newcommand{\denselist}{\itemsep 0pt\partopsep 0pt}

\title{Stochastic Planning and Lifted Inference}

\author{
Roni Khardon \\Department of Computer Science \\ Tufts University \\ {\tt roni@cs.tufts.edu}
\and
Scott Sanner \\ Department of Industrial Engineering \\ University of Toronto\\ {\tt ssanner@mie.utoronto.ca}}

\begin{document}

\maketitle

\begin{abstract}
  Lifted probabilistic inference (Poole, 2003) and symbolic dynamic programming for lifted 
  stochastic planning (Boutilier et al, 2001) were introduced around the same time as algorithmic efforts to use abstraction in  
  stochastic systems.
  Over the years, 
  these ideas
  evolved into two distinct lines of research, each supported by
  a rich literature.
  Lifted probabilistic inference focused
  on efficient arithmetic operations on template-based graphical
  models under a finite domain assumption while symbolic dynamic
  programming focused on supporting sequential decision-making in rich
  quantified logical action models and on open domain reasoning.  Given their
  common motivation but 
  different focal points, both lines of research have yielded 
  highly complementary innovations.  In this chapter, we aim to help close
  the gap between these two research areas by providing an overview of
  lifted stochastic planning from the perspective of probabilistic inference, 
  showing strong connections to other chapters in this book.
  This also allows us to define {\em generalized lifted inference} as a paradigm 
  that unifies these areas
  and elucidates open problems for future research that can benefit
  both lifted inference and stochastic planning.
\end{abstract}

\section{Introduction}

In this chapter we illustrate that stochastic planning can be viewed
as a specific form of probabilistic inference and show that recent symbolic
dynamic programming (SDP) algorithms for the planning problem can be
seen to perform  ``generalized lifted
inference'', thus making a strong connection to other chapters in this
book.  As we discuss below, although the SDP formulation is more
expressive in principle, work on SDP to date has largely focused on
algorithmic aspects of reasoning in open domain models with rich
quantified logical structure whereas lifted inference has largely
focused on aspects of efficient arithmetic computations
over finite domain (quantifier free) template-based models.
The contributions in these areas are therefore largely along different dimensions.
However, the intrinsic relationships between these problems
suggest a strong opportunity for cross-fertilization where the
true scope of generalized lifted inference can be achieved.  This
chapter intends to highlight these relationships and 
lay out a paradigm for generalized lifted inference that subsumes both
fields and offers interesting opportunities for future research.

To make the discussion concrete, let us introduce a running example for stochastic
planning and the kind of generalized solutions that can be achieved.
For illustrative purposes, we borrow a planning domain from Boutilier
et.\ al.~\cite{BoutilierRePr01} that we refer to as \textsc{BoxWorld}.
In this domain, outlined in Figure~\ref{fig:boxworld}, there are
several cities such as $\london$, $\paris$ etc., trucks $\truck_1$,
$\truck_2$ etc., and boxes $\tbox_1$, $\tbox_2$ etc. The agent can load
a box onto a truck or unload it and can drive a truck from one city to
another.  When any box 
has been delivered to a
specific city, $\paris$, the agent receives a positive
reward.  The agent's planning task is to find a policy for action
selection that maximizes this reward over some planning horizon.

\begin{figure}[t!]

\caption{A formal desciption of the \textsc{BoxWorld} adapted
  from~\cite{BoutilierRePr01}. We use a simple STRIPS-like~\cite{strips}
  add and delete list representation of actions and, as a simple
  probabilistic extension in the spirit of PSTRIPS~\cite{pso}, we
  assign probabilities that an action successfully executes
  conditioned on various state properties.\label{fig:boxworld}}
\vspace{2mm}
\begin{center}
\fbox{\begin{minipage}{\columnwidth}
{
\centering \small 
\begin{itemize} 
\item {\it Domain Object Types (i.e., sorts)}: $\TBox$, $\Truck$, $\City = \{ \paris, \ldots \}$
\item {\it Relations (with parameter sorts)}: \\BoxIn: $\BIn(\TBox,\City)$, TruckIn: $\TIn(\Truck,\City)$, BoxOn: $\On(\TBox,\Truck)$
\item {\it Reward}: if $\exists B, \BIn(B,\paris)$ then 10 else 0
\item {\it Actions (with parameter sorts)}:
  \begin{itemize} \denselist
  \item $\load(\TBox:B,\Truck:T,\City:C)$: %
    \begin{itemize} \denselist
    \item Success Probability: if $(\BIn(B,C) \wedge \TIn(T,C))$ then .9 else 0
    \item Add Effects on Success: $\{ \On(B,T) \}$
    \item Delete Effects on Success: $\{ \BIn(B,C) \}$
    \end{itemize}
  \item $\unload(\TBox:B,\Truck:T,\City:C)$: %
    \begin{itemize} \denselist
    \item Success Probability: if $(\On(B,T) \wedge \TIn(T,C))$ then .9 else 0
    \item Add Effects on Success: $\{ \BIn(B,C) \}$
    \item Delete Effects on Success: $\{ \On(B,T) \}$
    \end{itemize}
  \item $\drive(\Truck:T,\City:C_1,\City:C_2)$:
    \begin{itemize} \denselist
    \item Success Probability: if $(\TIn(T,C_1))$ then 1 else 0
    \item Add Effects on Success: $\{ \TIn(T,C_2) \}$
    \item Delete Effects on Success: $\{ \TIn(T,C_1) \}$
    \end{itemize}
  \item $noop$
    \begin{itemize} \denselist
    \item Success Probability: 1
    \item Add Effects on Success: $\emptyset$
    \item Delete Effects on Success: $\emptyset$
    \end{itemize}
  \end{itemize}
\end{itemize}}
\end{minipage}}
\end{center}
\end{figure}

Our objective in lifted stochastic planning is to obtain an
abstract policy, for example, like the one shown in Figure~\ref{fig:vfun_and_policy}.
In order to get some box to $\paris$, the agent should drive a truck
to the city where the box is located, load the box on the truck, drive
the truck to $\paris$, and finally unload the box in $\paris$.  This
is essentially encoded in the symbolic value function shown in
Fig.~\ref{fig:vfun_and_policy}, which was computed by discounting rewards
$t$ time steps into the future by $0.9^t$.

Similar to this example,
for some problems we can obtain a solution which is
described abstractly and is independent of the specific problem
instance or even its size --- for our example problem the description of
the solution does not depend on the number of cities, trucks or boxes, or
on  knowledge of the particular location of any specific truck.
Accordingly, one might hope that computing such a solution can be done without knowledge of these quantities and in time complexity independent of them.
This is the computational advantage of symbolic stochastic planning
which we associate with lifted inference in this chapter. 

The next two subsections expand on the connection between planning and inference, identify opportunities for lifted inference, and use these observations to define a new setup which we call {\em generalized lifted inference} which abstracts some of the work in both areas and provides new challenges for future work.

\begin{figure}[t!]
\caption{A decision-list representation of the optimal policy and 
  expected discounted reward for the \textsc{BoxWorld} problem.
  The optimal action parameters in the \emph{then} conditions correspond to the existential
  bindings that made the \emph{if} conditions true.
  \label{fig:vfun_and_policy}
}
\vspace{2mm}
\begin{center}
\fbox{\begin{minipage}{\columnwidth}
{
 \small
if $(\exists B, \BIn(B,\paris))$ then do $\noop$ (value = 100.00)\\
else if $(\exists B,T, \TIn(T,\paris) \wedge \On(B,T))$ then do $\unload(B,T,\paris)$ (value = 89.0)\\
else if $(\exists B,C,T, \On(B,T) \wedge \TIn(T,C))$ then do $\drive(T,C,\paris)$ (value = 80.0)\\
else if $(\exists B,C,T, \BIn(B,C) \wedge \TIn(T,C))$  then do $\load(B,T,C)$ (value = 72.0)\\
else if $(\exists B,C_1,T,C_2, \BIn(B,C_1) \wedge \TIn(T,C_2))$ then do $\drive(T,C_2,C_1)$ (value = 64.7)\\
else do $\noop$ (value = 0.0)
}
\end{minipage}}
\end{center}\end{figure}

\subsection{Stochastic Planning and Inference}

Planning is the task of choosing what actions to take to achieve some
goals or maximize long-term reward.  When the dynamics of the world are 
deterministic, that is, each action has exactly one known outcome,
then the problem can be solved through logical inference. That is,
inference rules can be used to deduce the outcome of individual actions given the
current state, and by combining inference steps one can prove that the
goal is achieved. In this manner a proof of goal achievement embeds a
plan. This correspondence was at the heart of McCarthy's
seminal paper \cite{McCarthy58} that introduced the topic of AI and viewed
planning as symbolic logical inference.  Since this formulation uses
first-order logic, or the closely related situation calculus, lifted
logical inference can be used to solve deterministic planning
problems.

When the dynamics of the world are non-deterministic, this
relationship is more complex.  In particular, in this chapter we focus
on the stochastic planning problem where an action can have multiple
possible known outcomes that occur with known state-dependent
probabilities.  Inference in this case must reason about probabilities
over an exponential number of state trajectories for some
planning horizon.
While lifted inference and planning may seem to be entirely different
problems, analogies have been made between the two fields in several
forms~\cite{Attias03,ToussaintSt06,domshlak2006,LangTo09,FurmstonB10,LiuI12,ChengLCI13,LeeMaDe14,LeeMaDe16,issakkimuthu2015hop,MeentPTW16}.
To make the connections concrete, consider a finite domain and the finite
horizon goal-oriented version of the \textsc{BoxWorld} planning
problem of Figure~\ref{fig:boxworld}, e.g., two boxes, three trucks,
and four cities and a planning horizon of 10 steps where the goal is
to get some box in $\paris$.  In this case, 
the value of a state, $V(S)$, corresponds to the probability of
achieving the goal, and
goal achievement can be
modeled as a specific form of inference in a Bayesian network or
influence diagram.  

\begin{figure*}
\begin{center}
\fbox{
\begin{minipage}{2.1in}
\begin{center}
{\includegraphics[height=1.1in,angle=0]{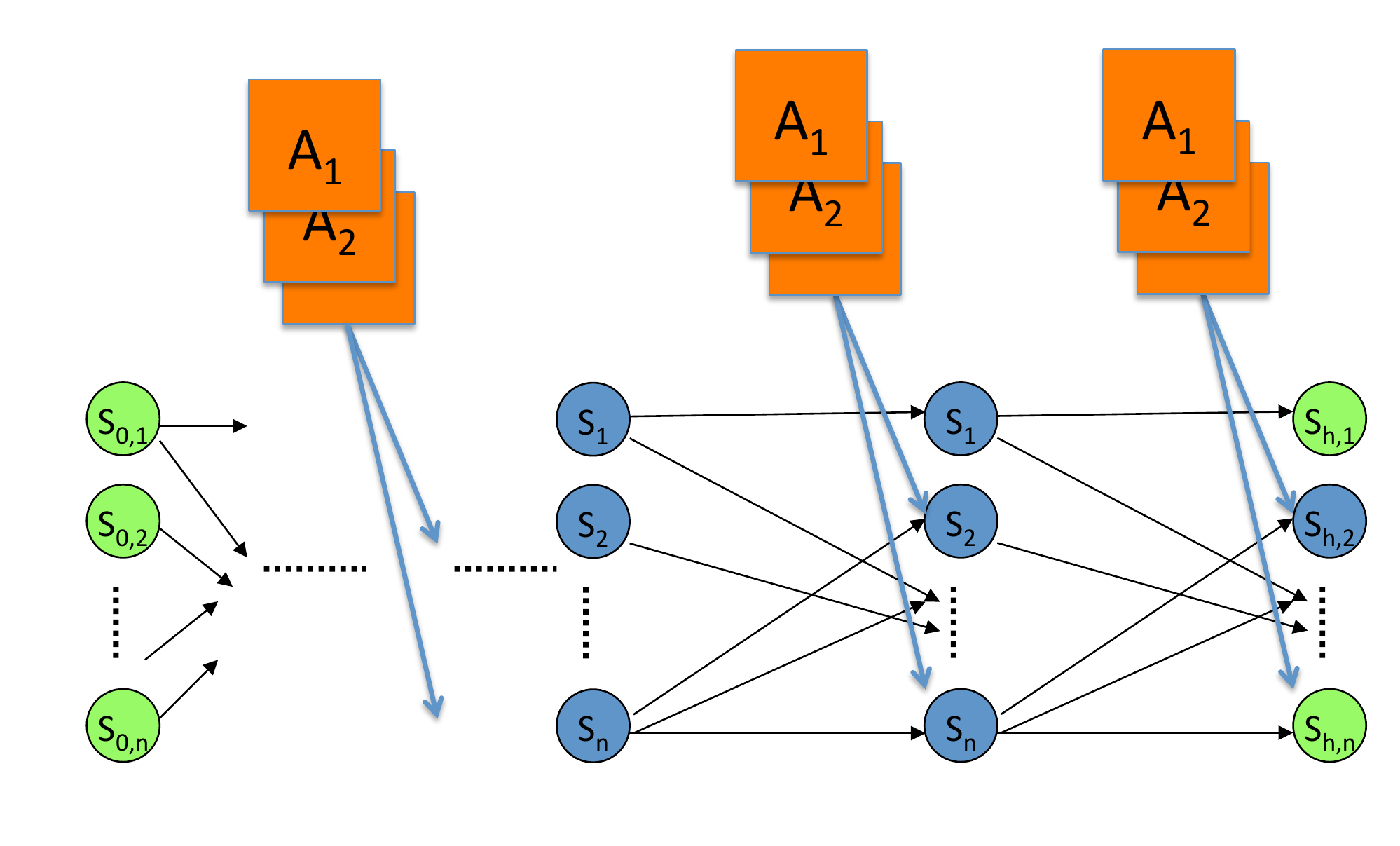}}

(a)
\end{center}
\end{minipage}
}
\fbox{
\begin{minipage}{2.1in}
\begin{center}
{\includegraphics[height=1.1in,angle=0]{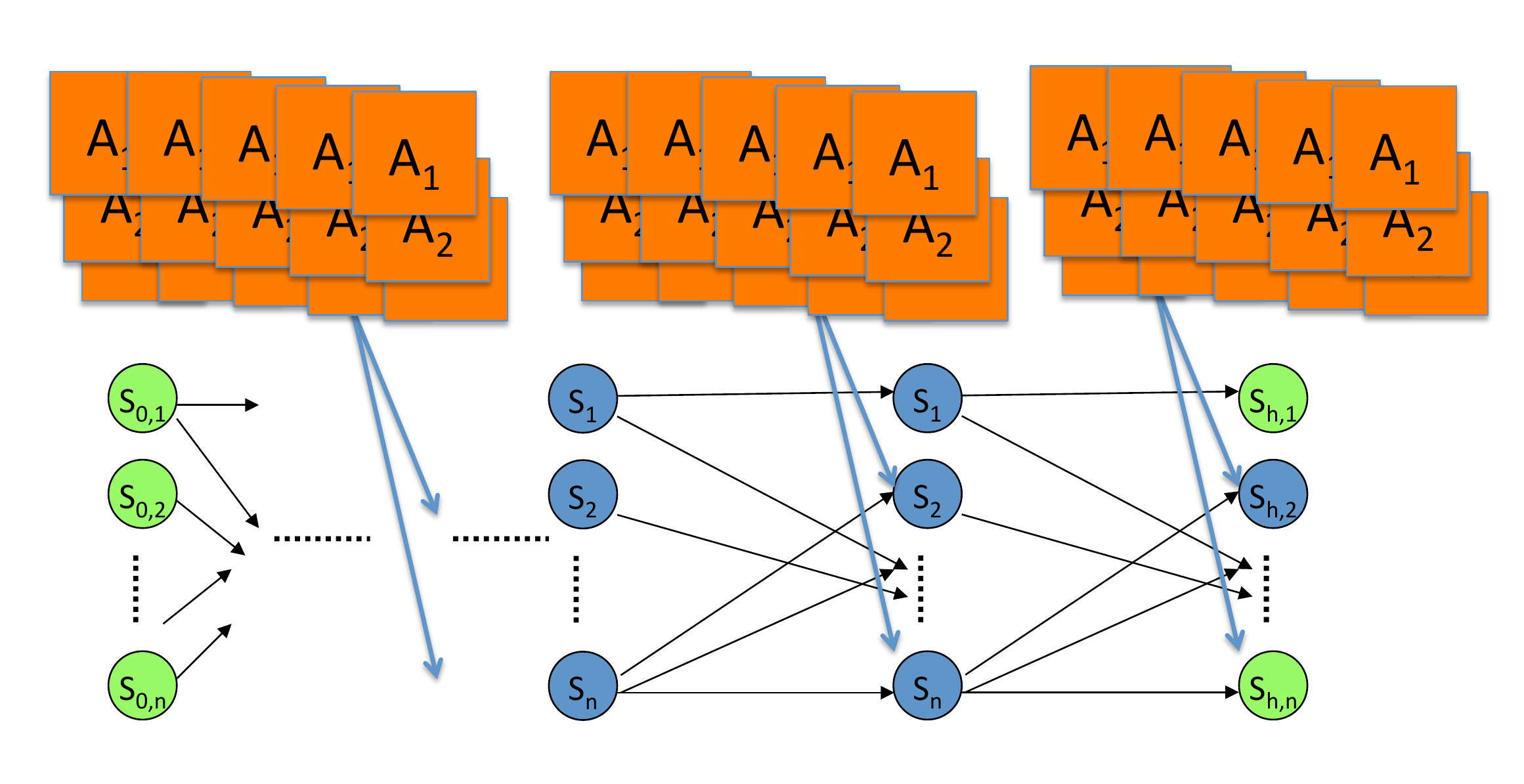}}

(b)
\end{center}
\end{minipage}
}
\end{center}
\caption{Planning as inference: conditioning on start and goal state.
(a) Conformant planning -- actions selected per time step without knowledge of the state.
(b) An exponential size policy at each time step determines action selection. The transition depends on the current state and policy's actions for that state.
}
\label{fig:PasI}
\end{figure*}

We start by considering the {\em conformant planning problem} where 
the intended solution is an explicit sequence of actions.
In this case, the sequence of actions is determined in advance and 
action choice at the $i$th step does not depend on the actual state at the $i$th step.
For this formulation,
one can build a Dynamic Bayesian Network (DBN) model where
each time slice represents the state at that time and action nodes
affect the state at the next time step, as in
Figure~\ref{fig:PasI}(a).  The edges in this diagram capture
$p(S'|S,A)$, where $S$ is the current state, $A$ is the current action
and $S'$ is the next state, and each of $S,S',A$ is represented by
multiple nodes to show that they are given by a collection of
predicates and their values.  Note that, since the world dynamics are
known, the conditional probabilities for all nodes in the graph
are known. 
As a result, 
the goal-based planning problem where a goal $G$ must hold at the last step,
can be modeled using standard
inference. The value of conformant planning is given by marginal MAP (where
we seek a MAP value for some variables but take expectation over the remaining variables)
\cite{domshlak2006,LeeMaDe14,LeeMaDe16}:
\begin{eqnarray*}
V_{\mbox{\footnotesize{conformant}}}(S_0) 
& = & \max_{A_0}, \ldots, \max_{A_{N-1}} 
Pr( G | S_0, A_0, \ldots, A_{N-1})  \\
& = & \max_{A_0}, \ldots, \max_{A_{N-1}} 
\sum_{S_1,S_2,\ldots,S_{N}} 
Pr( G,S_1,\ldots,S_N | S_0, A_0, \ldots, A_{N-1}). 
\end{eqnarray*}
The optimal conformant plan is extracted using
argmax instead of max in the equation.

The standard MDP formulation with a reward per time time step which is accumulated can be handled similarly, by normalizing the cumulative reward and adding a binary node $G$ whose probability of being true is a function of the normalized cumulative reward. Several alternative formulations of planning as inference have been proposed by defining an auxiliary distribution over finite trajectories which captures utility weighted probability distribution over the trajectories \cite{ToussaintSt06,FurmstonB10,LiuI12,ChengLCI13,MeentPTW16}. While the details vary, the common theme among these approaches is that the planning objective is equivalent to calculating the partition function (or ``probability of evidence") in the resulting distribution. 
This achieves the same effect as adding a node $G$ that depends on the cumulative reward.
To simplify the discussion, we continue the presentation with the simple goal based formulation.

The same problem can be viewed from a Bayesian perspective, treating actions as random variables with an uninformative prior. In this case we can use
\[
Pr(G | S_0, A_0, \ldots, A_{N-1}) = 
\frac
{Pr(A_0, \ldots, A_{N-1}| G, S_0 ) Pr(G | S_0)}{Pr(A_0, \ldots, A_{N-1}|  S_0)}.
\]
to observe that  \cite{Attias03,ToussaintSt06,LangTo09}
\[
\argmax_{A_0}, \ldots, \max_{A_{N-1}}  Pr(G | S_0, A_0, \ldots, A_{N-1}) = \argmax_{A_0}, \ldots, \max_{A_{N-1}}  Pr(A_0, \ldots, A_{N-1}| G, S_0 )
\]
and therefore one can alternatively maximize the probability conditioned on $G$.

However, linear plans, as the ones produced by the conformant setting, are not optimal for probabilistic planning. In
particular, if we are to optimize goal achievement then we must allow
the actions to depend on the state they are taken in. That is, the
action in the second step is taken with knowledge of the probabilistic
outcome of the first action, which is not known in advance. 
We can achieve this by duplicating action nodes, with a copy for each possible value of the state variables, as illustrated in 
Figure~\ref{fig:PasI}(b). This represents a separate policy associated with each horizon depth which is required because finite horizon problems have non-stationary optimal policies.
In this case, state transitions depend on the identity of the current state and the action variables associated with that state. 
The corresponding inference problem can
be written as follows:
\begin{equation}
\label{eq:VoptV1}
V(S_0) = \max_{A_0(S_0)}, \ldots, \max_{A_{N-1}(S_N-1)} Pr(G | S_0, A_0(S_0), \ldots, A_{N-1} (S_N-1 ) ). 
\end{equation}
However, the number of random variables in this formulation is
prohibitively large since we need the number of original action variables to be multiplied by the size of the state space. 

Alternatively, the same desideratum, optimizing
actions with knowledge of the previous state, can be achieved without
duplicating variables in the equivalent formulation
\begin{align}
V(S_0) &  = 
\max_{A_0}
\sum_{S_1}
Pr(S_{1}|S_{0},A_{0})
\max_{A_1}
\sum_{S_2}
Pr(S_{2}|S_{1},A_{1})
\ldots
\nonumber
\\
& \ \ \ \ 
\max_{A_{N-2}}
\sum_{S_{N-1}}
Pr(S_{N-1}|S_{N-2},A_{N-2})
\max_{A_{N-1}}
\sum_{S_N}
Pr(S_N|S_{N-1},A_{N-1})
Pr(G|S_n). 
\label{eq:VoptV2}
\end{align}
In fact, this formulation is exactly the same as the finite horizon
application of the value iteration (VI) algorithm for (goal-based) Markov Decision
Processes (MDP) which is the standard formulation for sequential decision
making in stochastic environments.  The standard formulation abstracts
this by setting 
\begin{align}
V_0(S) & = Pr(G|S) 
\nonumber
\\
V_{k+1}(S) & =  
\max_{A}
\underbrace{\sum_{S'} Pr(S'|S,A) V_k(S')}_{Q(S,A)}.
\label{eq:VI} 
\end{align}
The optimal policy (at $S_0$) can be obtained as before by recording
the argmax values.  In terms of probabilistic inference, the problem
is no longer a marginal MAP problem because summation and maximization steps are
constrained in their interleaved order. But it can be seen as a natural extension of such inference questions with several alternating blocks of expectation and maximization.
We are not aware of an explicit study of such problems outside the planning context.

\subsection{Stochastic Planning and Generalized Lifted Inference}

Given that planning can be seen as an inference problem, one can try
to apply ideas of lifted inference to planning.  Taking the motivating
example from Figure~\ref{fig:boxworld}, let us specialize the reward
to a ground atomic goal $G$ equivalent to $\BIn(b^*,\paris)$ for
constants $b^*$ and $\paris$.  Then we can query
$\max_A Pr(\BIn(b^*,\paris)|s_0,A)$ to compute $V(S_0=s_0)$ where $s_0$ is
the concrete value of the current state.  

Given that
Figure~\ref{fig:boxworld} implies a complex relational specification
of the transition probabilities, lifted inference techniques are
especially well-placed to attempt to exploit the structure of this
query to perform inference in aggregate and thus avoid redundant
computations.
However, we emphasize that, even if lifted inference is used, this is a standard query in the graphical model where evidence constrains the value of some nodes, and the solution is a single number representing the corresponding probability (together with a MAP assignment to variables).

However, Eq~\ref{eq:VI} suggests an explicit additional structure for
the planning problem. In particular, the intermediate expressions $V_k(S)$
include the values (the probability of reaching the goal in $k$ steps)
for all possible concrete values of $S$. Similarly, the final result
$V_N(S)$ includes the values for all possible start states. In addition, as in our running example we can consider more abstract rewards.
This suggests a
first generalization of the standard setup in lifted
inference. Instead of asking about a ground goal $Pr(\BIn(b^*,\paris))$ and expecting a single number as a response, we can abstract the setup in two ways: first, we can ask about
more general conditions such as $\exists B, Pr(\BIn(B,\paris))$ and second we can expect to get a structured result that specifies the
corresponding probability for every concrete state in the world.
If we had two box instances $b_1,b_2$ and $m$ truck instances $t_1,\ldots,t_m$, the answer for $V_1(S)$, i.e., the value for the goal based formulation with horizon one, might take the form:
\begin{center}
\vspace{-4mm}
\fbox{\begin{minipage}{\columnwidth}
{
if $(\BIn(b_1,\paris)\vee \BIn(b_2,\paris))$\\
$\mbox{\;}$ \hspace{10mm} then 
$V_1(S)=10$\\
else if $((\TIn(t_1,\paris) \!\wedge\! \On(b_1,t_1)) 
\!\vee\! 
\ldots \!\vee\! (\TIn(t_m,\paris) \!\wedge\! \On(b_2,t_m)))$\\
$\mbox{\;}$ \hspace{10mm} then 
$V_1(S)=9$\\
else 
$V_1(S)=0$.
}
\end{minipage}}
\end{center}
The significance of this is that the question can have a more general form and that the answer solves many problems simultaneously, providing the response as a case analysis depending on some properties of the state. 
We refer to this reasoning as {\em inference with
generalized queries and answers}.  In this context, the goal of lifted inference will
be to calculate a structured form of the reply directly. 

A second extension arises from the setup of generalized
queries. The standard form for lifted inference is to completely
specify the domain in advance. This means providing the number of
objects and their properties, and that the response to the query is calculated
only for this specific domain instantiation.
However, inspecting the solution in the previous paragraph it is obvious that we can at least hope to do better. The same solution can be described more compactly as 
\begin{center}
\vspace{-4mm}
\fbox{\begin{minipage}{\columnwidth}
{
if $(\exists B, \BIn(B,\paris))$\\
$\mbox{\;}$ \hspace{10mm} then 
$V_1(S)=10$\\
else if $(\exists B,\exists T, (\TIn(T,\paris) \!\wedge\! \On(B,T))$ \\
$\mbox{\;}$ \hspace{10mm} then 
$V_1(S)=9$\\
else 
$V_1(S)=0$.
}
\end{minipage}}
\end{center}
Arriving at such a solution requires us to 
allow open domain reasoning over all potential objects (rather than
grounding them, which is impossible in open domains), and to extend 
ideas of lifted inference to exploit quantifiers and their structure. 
Following through with this idea, we can 
arrive at a \emph{domain-size independent value function and policy}
as the one shown in Figure~\ref{fig:vfun_and_policy}.  
In this context, the goal of lifted inference will
be to calculate an abstracted form of the reply directly. 
We call this problem {\em inference with generalized models}.
As we describe in this chapter, SDP algorithms are able to perform this type of inference.

The previous example had enough structure and a special query that allowed the solution to be specified without any knowledge of the concrete problem instance. This property is not always possible.  
For example, consider a setting 
where we get one unit of reward for every box in $\paris$: 
$\sum_{B:\TBox} \mbox{[if $(\BIn(B,\paris))$ then 1 else 0]}$.
In addition, consider the case where, after the agent takes their action, any box which is not on a truck disappears with probability $0.2$.
In this case, we can still potentially calculate an abstract solution, but it requires access to more complex properties of the state, and in some cases the domain size (number of objects) in the state. For our example this gives:
\begin{center}
\vspace{-4mm}
\fbox{\begin{minipage}{\columnwidth}
{
Let $n=(\#_B, \BIn(B,\paris))$\\
if $(\exists B,\exists T, (\TIn(T,\paris) \!\wedge\! \On(B,T))$ \\
$\mbox{\;}$ \hspace{10mm} then 
$V_1(S)=n*8+ 7.2$\\
else 
$V_1(S)=n*8$.
}
\end{minipage}}
\end{center}
Here we have introduced a new notation for count expressions where, for example, $(\#_B, \BIn(B,\paris))$ counts the number of boxes in Paris in the current state. 
To see this result note that any existing box in Paris disappears 20\% of the time and that
a box on a truck is successfully unloaded 90\% of the time but remains and does not disappear only in 80\% of possible futures leading to the value 7.2.
This is reminiscent of the type of expressions that arise in existing lifted inference problems and solutions. 
Typical solutions to such problems
involve parameterized expressions over the domain (e.g., counting, summation, etc.),
and critically do not always require closed-domain reasoning (e.g., \textit{a priori}
knowledge of the number of boxes).  They are therefore suitable for inference with generalized models.
Some work on SDP has approached lifted inference for problems with this level of complexity, including exogenous activities (the disappearing boxes) and additive rewards.
But, as we describe in more detail, the solutions for these cases are much less well understood and developed.

To recap, our example illustrates that stochastic planning potentially enables abstract solutions that 
might be amenable to lifted computations. 
SDP solutions for planning problems have
focused on the computational advantages arising from these
expressive generalizations. At the same time, the focus in SDP algorithms has largely been on
problems where the solution is completely independent of domain size and does not require numerical properties of the state. 
These algorithms have thus skirted some of
the computational issues that are typically tackled in lifted inference. It is the
combination of these aspects, as illustrated in the last example, which we call {\bf generalized lifted
inference}. As the discussion suggests, generalized lifted inference is still very much an open problem. In addition to providing a survey of existing SDP algorithms, the goal of this chapter is to highlight the opportunities and challenges in this exciting area of research.

\section{Preliminaries}

This section provides a formal description of the representation language, the relational planning problem, and the description of the running example in this context.

\subsection{Relational Expressions and their Calculus of Operations}

The computation of SDP algorithms is facilitated by a representation
that enables compact specification of functions over world
states. Several such representations have been devised and used. In
this chapter we chose to abstract away some of those details and focus
on a simple language of relational expressions. This is closest to the
GFODD representation of \cite{JoshiKeKh11,JoshiKhRaTaFe13}, but it resembles the case
notation of \cite{BoutilierRePr01,SannerBo09}.

\subfour{Syntax}
We assume familiarity with basic concepts and notation in  first order logic (FOL) \cite{Lloyd87,RussellNo95,ChangKe90}. 
Relational expressions are similar to expressions in 
FOL. They are defined relative to a relational signature, with a
finite set of predicates $p_1, p_2, \ldots, p_n$ each with an
associated arity (number of arguments), a countable set of variables
$x_1, x_2, \ldots$, and a set of constants $c_1, c_2, \ldots, c_m$. We
do not allow function symbols other than constants (that is, functions
with arity $\geq 1$).  
A term is a
variable (often denoted in uppercase) or constant (often denoted in lowercase)
and an atom is either an equality between two
terms or a predicate with an appropriate list of terms as arguments.
Intuitively, a term refers to an object in the world of interest and
an atom is a property which is either true or false.

We illustrate relational expressions informally by some examples. In
FOL we can consider open formulas that have unbound variables. For
example, the atom $color(X,Y)$ is such a formula and its truth value
depends on the assignment of $X$ and $Y$ to objects in the world.  To
simplify the discussion, we assume for this example that arguments are
typed (or sorted) and $X$ ranges over ``objects'' and $Y$ over ``colors''.  We can
then quantify over these variables to get a sentence which will be
evaluated to a truth value in any concrete possible world. For
example, we can write $[\exists Y, \forall X, color(X,Y)]$ expressing
the statement that there is a color associated with all objects.
Generalized expressions allow for more general open formulas that
evaluate to numerical values.  For example, $E_1=[\mbox{if }
  color(X,Y) \mbox{ then 1 else 0}]$ is similar to the previous logical
expression but $E_2 =[\mbox{if } color(X,Y) \mbox{ then 0.3 else
    0.5}]$ returns non-binary values.

Quantifiers from logic are replaced with aggregation operators that
combine numerical values and provide a generalization of the logical
constructs. In particular, when the open formula is restricted to
values 0 and 1, the operators $\max$ and $\min$ simulate existential
and universal quantification.  Thus, $[\max_{Y}, \min_{X}, \mbox{if }
  color(X,Y) \mbox{ then 1 else 0}]$ is equivalent to the logical
sentence $[\exists Y, \forall X, color(X,Y)]$ given above.  But we can
allow for other types of aggregations. For example, $[\max_{Y},
  \mbox{sum}_{X}, \mbox{if }$ $color(X,Y)$ $\mbox{ then 1 else 0}]$
evaluates to the largest number of objects associated with one color,
and the expression $[\mbox{sum}_{X}, \min_{Y},$ $\mbox{if }
  color(X,Y)$ $\mbox{ then 0 else 1}]$ evaluates to the number of
objects that have no color association.
In this
manner, a generalized expression represents a function from possible
worlds to numerical values and, as illustrated, can capture interesting properties of the state.

Relational expressions are also related to work in statistical
relational learning \cite{RichardsonDo06,Problog,LiftedWMC}.  For example, if the
open expression $E_2$ given above captures probability of ground facts for the
predicate $color()$ and the ground facts are mutually independent then
$[\mbox{product}_{X}, \mbox{product}_{Y}, \mbox{if } color(X,Y)$ $\mbox{
    then 0.3 else 0.5}]$ captures the joint probability for all facts
for $color()$. Of course, the open formulas in logic can include more
than one atom and similarly expressions can be more involved. 

In the following we will drop the cumbersome if-then-else notation and
instead will assume a simpler notation with a set of mutually exclusive conditions which we refer to as {\em cases}.  In particular, an
expression includes a set of mutually exclusive open formulas in FOL
(without any quantifiers or aggregators) 
denoted $c_1,\ldots,c_k$ associated with corresponding numerical values
$v_1,\ldots,v_k$.  The list of cases refers to a finite set of
variables $X_1,\ldots,X_m$. A generalized expression is given by a
list of aggregation operators and their variables and the list of
cases $[agg_{X_1}, agg_{X_2}, \ldots , agg_{X_m}
  [c_1:v_1,\ldots,c_k:v_k]]$ so that the last expression is
canonically represented as $[\mbox{product}_{X}, \mbox{product}_{Y},
    [color(X,Y):0.3; \neg color(X,Y):0.5]]$.

\subfour{Semantics}
The semantics of expressions is defined inductively exactly as in
first order logic and we skip the formal definition.  
As usual, an expression is evaluated in an \emph{interpretation}  also known as a possible world. 
In our context, an interpretation specifies (1) a
finite set of $n$ domain elements also known as objects, (2) a mapping
of constants to domain elements, and (3) the truth values of all the
predicates over tuples of domain elements of appropriate size to match
the arity of the predicate.
Now,
given an expression $B=(agg_X,\ f(X))$, an interpretation $I$, and a
substitution $\zeta$ of variables in $X$ to objects in $I$, one can
identify the case $c_i$ which is true for this substitution.  Exactly
one such case exists since the cases are mutually exclusive and exhaustive.
Therefore, the value associated with $\zeta$ is $v_i$.  These values
are then aggregated using the aggregation operators.  For example,
consider again the expression $[\mbox{product}_{X},
    \mbox{product}_{Y}, [color(X,Y):0.3; \neg color(X,Y):0.5]]$ and an
    interpretation $I$ with objects $a,b$ and where $a$ is associated
    with colors black and white and $b$ is associated with color
    black.  In this case we have exactly 4 substitutions evaluating to
    0.3, 0.3, 0.5, 0.3. Then the final value is $0.3^3 \cdot 0.5$.

\subfour{Operations over expressions}
Any binary operation $op$ over real values can be generalized to open
and closed expressions in a natural way. If $f_1$ and $f_2$ are two
closed expressions, $f_1\ op\ f_2$ represents the function which maps
each interpretation $w$ to $f_1(w)\ op\ f_2(w)$.
This provides a definition but not an implementation of binary
operations over expressions.  
For implementation,
the work in \cite{JoshiKeKh11} showed that if the binary operation is
{\em safe}, i.e.,\ it distributes with respect to all aggregation
operators, then there is a simple algorithm (the Apply procedure)
implementing the binary operation over expressions.  For example, $+$
is safe w.r.t.\ $\max$ aggregation, and it is easy to see that
$(\max_X f(X)) + (\max_X g(X))$ = $\max_X \max_Y f(X)+ g(Y)$, and the
open formula portion of the result can be calculated directly from the
open expressions $f(X)$ and $g(Y)$.  
Note that we need to standardize
the expressions apart, as in the renaming of $g(X)$ to $g(Y)$ for such
operations. 
When $f(x)$ and $g(y)$
are open relational expressions
the result can be computed through a cross product of the cases. 
For example,
\begin{align*}
[\max_{X}, \min_{Y} \, [color & (X,Y) :3; \neg color(X,Y):5]] \; \oplus \;
[\max_{X}, [box(X):1; \neg box(X):2]] 
\\
= [\max_Z, \max_{X}, \min_{Y} \, [& color(X,Y)\wedge box(Z):4; \neg color(X,Y)\wedge box(Z):6; 
\\
& color(X,Y)\wedge \neg box(Z):5; \neg color(X,Y)\wedge \neg box(Z):7]]
\end{align*}
When the binary operation is not safe then this procedure
fails, but in some cases, operation-specific algorithms can be
used for such combinations.\footnote{For example, a product of expressions that include only product aggregations, which is not safe, can be obtained by scaling the result with a number that depends on domain size, and 
$[\prod_{x_1} \prod_{x_2} \prod_{x_3} f(x_1,x_2,x_3)] 
\otimes
[\prod_{y_1} \prod_{y_2} g(y_1,y_2)]$ is euqal to 
$
[\prod_{x_1} \prod_{x_2} \prod_{x_3} 
[f(x_1,x_2,x_3)
\times g(x_1,x_2)^{1/n} ] ]
$ when the domain has $n$ objects.
}

As will become clear later, to implement SDP we need the binary
operations $\oplus$, $\otimes$, $\max$ and the aggregation includes
$\max$ in addition to aggregation in the reward function.  Since
$\oplus$, $\otimes$, $\max$ are safe with respect to $\max,\min$
aggregation one can provide a complete solution 
when the reward is restricted to have $\max,\min$ aggregation. 
When this is not the case, for example when using sum aggregation in the
reward function,  one requires a special algorithm for the
combination. Further details are provided in \cite{JoshiKeKh11,JoshiKhRaTaFe13}.

\subfour{Summary}
Relational expressions are closest to the GFODD representation of
\cite{JoshiKeKh11,JoshiKhRaTaFe13}.  Every case $c_i$ in a relational expression corresponds to a path or set of paths in the GFODD, all of which reach the same leaf in the graphical representation
of the GFODD.  GFODDs are potentially more compact than relational expressions since paths share common subexpressions, which can lead to an exponential reduction in size. On the other hand, GFODDs require special algorithms for their manipulation.
Relational expressions are also similar to the
case notation
of~\cite{BoutilierRePr01,SannerBo09}. However, in contrast with that representation, cases are not allowed to include any quantifiers and instead quantifiers and general aggregators are globally applied over the cases, as in standard quantified normal form in logic.

\subsection{Relational MDPs}

In this section we define MDPs, starting 
with the basic case with enumerated state and action spaces,
and then providing the relational representation.

\subfour{MDP Preliminaries}
We assume familiarity with basic notions of Markov Decision Processes
(MDPs) \cite{RussellNo09,Puterman1994}.  Briefly,
a MDP is a tuple $\langle S,A,P,R,\gamma \rangle$ given by a set of
states $S$, set of actions $A$, transition probability $Pr(S'|S,A)$, immediate
reward function $R(S)$
and discount factor $\gamma<1$.  The solution of a MDP is a policy
$\pi$
that maximizes the expected discounted total reward
obtained by following that policy starting from any state.  The Value
Iteration algorithm (VI) informally introduced in Eq~\ref{eq:VI}, calculates the
optimal value function by iteratively performing Bellman backups,
$V_{k+1} = T[V_k]$, defined for each state $s \in S$ as,
\begin{equation}
\label{eq:viflat}
V_{k+1}(s) = T[V_k](s) \leftarrow \max_{a \in A} \{ R(s) + \gamma \sum_{s' \in S} Pr(s'|s,a) V_k(s')\}.
\end{equation}
Unlike Eq~\ref{eq:VI}, which was goal-oriented and had only a single
reward at the terminal horizon, here we allow the reward R(S) to accumulate
at all time steps as typically allowed in MDPs.  
If we iterate the update until convergence, we get the
optimal infinite horizon value function typically denoted by $V^*$ and optimal stationary policy $\pi^*$.
For finite horizon problems, which is the topic of this chapter, we simply stop the iterations at a
specific $k$. 
In general, the optimal policy for the finite horizon case is not stationary, that is, we might make different choice in the same state depending on how close we are to the horizon. 

\subfour{Logical Notation for Relational MDPs (RMDPs)}  
RMDPs are simply MDPs where the states and actions are
described in a function-free first order logical language. 
A state corresponds to an interpretation over the corresponding logical signature, and actions are transitions between such interpretations.

A relational planning problem is specified by providing the logical
signature, the start state, the transitions as controlled by actions,
and the reward function.  As mentioned above, one of the advantages of
relational SDP algorithms is that they are intended to produce an
abstracted form of the value function and policy that does not require
specifying the start state or even
the number of objects $n$ in the interpretation at planning
time.  This yields policies that generalize across domain sizes.  
We therefore need to explain how one can use logical notation to represent the
transition model and reward function in a manner that does not depend on domain size. 

Two types of transition models have been considered in the literature:
\begin{itemize}
\item {\bf Endogenous Branching Transitions:} In the basic form, state transitions
  have limited stochastic branching due to a finite number of action
  outcomes.  The agent has a set of action types $\{A\}$ each
  parametrized with a tuple of objects to yield an action template
  $A(X)$ and a concrete ground action $A(x)$ (e.g. template
  $\unload(B,T)$ and concrete action
  $\unload(\mathit{box23},\mathit{truck1})$). 
  Each agent action has a finite number of action
  variants $A_j(X)$ (e.g., action success vs. action failure), and
  when the user performs $A(X)$ in state $s$ one of the variants is
  chosen randomly using the state-dependent action choice distribution
  $Pr(A_j(X) | A(X))$.  
    To simplify the presentation we follow
\cite{WangJoKh08,JoshiKeKh11} and require that $Pr(A_j(X)|A(X))$ are given by open expressions, i.e., they have no aggregations and cannot introduce new
variables.  For example, in \textsc{BoxWorld}, the agent
  action $\unload(B,T,C)$ has success outcome $\unloadS(B,T,C)$ and
  failure outcome $\unloadF(B,T,C)$ with action outcome distribution
  as follows:
\begin{align}
  P(\unloadS(B,T,C) | \unload(B,T,C)) & = [(\On(B,T) \wedge \TIn(T,C)): .9; \neg: 0] \nonumber \\
  P(\unloadF(B,T,C) | \unload(B,T,C)) & = [(\On(B,T) \wedge \TIn(T,C)): .1; \neg: 1]
   \label{eq:stoch_act_ex} 
\end{align}
where, to simplify the notation, the last case is shortened as $\neg$ to denote that it complements previous cases.
This provides the distribution over deterministic outcomes of 
actions.

The deterministic action dynamics are specified by providing an open expression,
capturing successor state axioms~\cite{reiter_KIA}, for each variant
$A_j(X)$ and predicate template $p'(Y)$. Following \cite{WangJoKh08} we
call these expressions TVDs, standing for truth value diagrams.  The corresponding TVD,
$T(A_j(X),p'(Y))$, is an open expression that specifies the truth value
of $p'(Y)$ {\em in the next state} 
  (following standard practice we use prime to denote that the predicate refers to the next state) when $A_j(X)$ has been executed {\em
  in the current state}.  
The arguments $X$ and $Y$ are intentionally different logical variables as this allows us to specify the truth value of all instances of $p'(Y)$ simultaneously.  
Similar to the choice probabilities we follow
\cite{WangJoKh08,JoshiKeKh11} and assume that 
  TVDs $T(A_j(X),p'(Y))$ have no aggregations and cannot introduce new
variables.
This implies that the regression and
product terms in the SDP algorithm of the next section do not change the aggregation
function, thereby enabling analysis of the algorithm.
Continuing our \textsc{BoxWorld} example, we define the TVD for $\BIn'(B,C)$ for
$\unloadS(B_1,T_1,C_1)$ and $\unloadF(B_1,T_1,C_1)$ as follows:
\begin{align}
  \BIn'(B,C) \equiv & T(\unloadS(B_1,T_1,C_1),\BIn'(B,C)) \nonumber \\
  \equiv & [(\BIn(B,C) \lor \nonumber \\
  & \, ((B_1=B)\land (C_1=C)  \land \On(B_1,T_1) \land \TIn(T_1,C_1))):1; \neg: 0] \nonumber \\
 & \nonumber  \\
  \BIn'(B,C) \equiv & T(\unloadF(B_1,T_1,C_1),\BIn'(B,C)) \nonumber \\
  \equiv & [\BIn(B,C):1; \neg:0] 
  \label{eq:ssa_ex}
\end{align}
Note that each TVD has exactly two cases, one leading to the outcome 1 and the other leading to the outcome 0.
Our algorithm below will use these cases individually.
Here we remark that since the next state (primed) only depends on the previous
state (unprimed), we are effectively logically encoding the Markov assumption of MDPs.
\item {\bf Exogenous Branching Transitions:} The more complex form combines the
  endogenous model with an exogenous stochastic process that affects
  ground atoms independently.  As a simple example in our
  \textsc{BoxWorld} domain, we might imagine that with some small
  probability, each box $B$ in a city $C$ ($\BIn(B,C)$) may
  independently randomly disappear (falsify $\BIn(B,C)$) owing to
  issues with theft or improper routing --- such an outcome is
  independent of the agent's own action.  
  Another more complicated example could be an
  inventory control problem where customer arrival at shops (and
  corresponding consumption of goods) follows an independent
  stochastic model.  Such exogenous transitions can be formalized
  in a number of ways~\cite{Sanner08,sanner:icaps07,JoshiKhRaTaFe13};
  we do not aim to commit to a particular representation in this chapter,
  but rather to mention its possibility and the computational
  consequences of such general representations.
\end{itemize}

Having completed our discussion of RMDP transitions, we now proceed to
define the reward $R(S,A)$, which can be any function of the state and
action, specified by a relational expression. 
Our running example with existentially quantified reward is given by
\begin{equation}
[\max_B [\BIn(B,\paris): 10; \neg  \BIn(B,\paris): 0]]
\label{eq:reward}
\end{equation}
but we will also consider additive reward as in 
\begin{equation}
[\sum_B [\BIn(B,\paris): 10; \neg  \BIn(B,\paris): 0]].
\label{eq:reward-additive}
\end{equation}

\section{Symbolic Dynamic Programming}

The SDP algorithm is a symbolic implementation of the value iteration algorithm. 
The algorithm repeatedly applies so-called decision-theoretic regression which is equivalent to one iteration of the value iteration algorithm.

As input to SDP we get closed
relational expressions for $V_k$ and $R$.  In addition, assuming that we
are using the \emph{Endogenous Branching Transition} model of the
previous section, we get open expressions for the probabilistic choice
of actions $Pr(A_j(X)|A(X))$ and for the dynamics of deterministic
action variants as TVDs. The corresponding expressions for the running example are given respectively in 
Eq~(\ref{eq:reward}), 
Eq~(\ref{eq:stoch_act_ex}) and Eq~(\ref{eq:ssa_ex}).

The following SDP algorithm of \cite{JoshiKeKh11} modifies the earlier
SDP algorithm of~\cite{BoutilierRePr01} and implements Eq~(\ref{eq:viflat}) using
the following 4 steps:
\begin{enumerate}
\item \label{sdp_1} {\bf Regression:} The $k$ step-to-go value
  function $V_k$ is regressed over every deterministic variant
  $A_j(X)$ of every action $A(X)$ to produce $\Regr(V_k, A_j(X))$.
Regression is conceptually similar to goal regression in
deterministic planning. That is, we identify conditions that need to occur before the action is taken in order to arrive at other conditions (for example the goal) after the action.  
However, here we need to regress all the conditions in the relational expression capturing the value function, so that we must regress 
each case $c_i$
of $V_k$ separately.  This can be done efficiently by replacing every atom in
each $c_i$ by its corresponding positive or negated portion of the TVD without changing the aggregation
function.  
Once this substitution is done, logical simplification (at the
  propositional level) can be used to compress the cases  by
  removing contradictory cases and simplifying the formulas. 
Applying this to regress $\unloadS(B_1,T_1,C_1)$ over the reward function given by Eq~(\ref{eq:reward}) we get:
\begin{align*}
  [\max_B \, [ & (\BIn(B,\paris) \lor \\
      & ((B_1=B)\land (C_1=\paris)  \land \On(B_1,T_1) \land \TIn(T_1,C_1))): 10; \neg: 0]]
\end{align*}
and regressing $\unloadF(B_1,T_1,C_1)$ yields
\begin{equation*}
[\max_B \, [\BIn(B,\paris): 10; \neg: 0]]
\end{equation*}
This illustrates the utility of compiling the transition model into the TVDs which allow for a simple implementation of deterministic regression.

\item \label{sdp_2} {\bf Add Action Variants:} The Q-function
  $Q_k^{A(X)}$ $=$ $R$ $\oplus$ $[\gamma$ $\otimes$
  $\oplus_j(Pr(A_j(X))$ $\otimes$ $Regr(V_k, A_j(X)))]$ for each
  action $A(X)$ is generated by combining regressed diagrams using the
  binary operations $\oplus$ and $\otimes$ over expressions.
  Recall that probability expressions do not refer to additional
  variables. The multiplication can therefore be done directly on the
  open formulas without changing the aggregation function.  As argued by
  \cite{WangJoKh08}, to guarantee correctness, both summation steps
  ($\oplus_j$ and $R\oplus$ steps) must standardize apart the functions
  before adding them.
  
  For our running example and assuming $\gamma=0.9$, we would need to compute the following:
  \begin{align*}
    Q_k&^{\unload(B_1,T_1,C_1)}(S)  = \\  R&(S) \oplus 0.9 \cdot \\
     [&(\Regr(V_0, \unloadS(B_1,T_1,C_1)) \otimes P(\unloadS(B_1,T_1,C_1) | \unload(B_1,T_1,C_1))) \oplus\\
      &(\Regr(V_0, \unloadF(B_1,T_1,C_1)) \otimes P(\unloadF(B_1,T_1,C_1) | \unload(B_1,T_1,C_1)))].
  \end{align*}
We next illustrate some of these steps. The multiplication by probability expressions can be done by cross product of cases and simplification. For $\unloadS$ this yields
\begin{align*}
  [\max_B \, [ & ((\BIn(B,\paris) \lor ((B_1=B)\land (C_1=\paris))) \\
               &   \land \On(B_1,T_1) \land \TIn(T_1,C_1)): 9; \neg: 0]]
\end{align*}
and for $\unloadF$ we get 
\begin{align*}
  [\max_B \, [ & \BIn(B,\paris)   \land (\On(B_1,T_1) \land \TIn(T_1,C_1)): 1;  \\
& \BIn(B,\paris) \land  \neg (\On(B_1,T_1) \land \TIn(T_1,C_1)): 10; \\
& \neg:  0]].
\end{align*}
Note that the values here are weighted by the probability of occurrence. For example the first case in the last equation has value 1=10*0.1 because when the preconditions of $\unload$ hold the variant $\unloadF$ occurs with $10\%$ probability. 
The addition of the last two equations requires standardizing them apart, performing the safe operation through cross product of cases, and simplifying. Skipping intermediate steps, this yields 
  \begin{align*}
[\max_B \, [ & \BIn(B,\paris):10;  \\
  & \neg \BIn(B,\paris) \land (B_1=B)\land (C_1=\paris)  \land \On(B_1,T_1) \land \TIn(T_1,C_1): 9; \\
  & \neg: 0]].
  \end{align*}
Multiplying by the discount factor scales the numbers in the last equation by 0.9 and finally standardizing apart and adding the reward and simplifying (again skipping intermediate steps) yields
  \begin{align*}
Q_0&^{\unload(B_1,T_1,C_1)}(S) = \\
[& \max_B \, [ \BIn(B,\paris):19;  \\
  & \qquad \, \neg \BIn(B,\paris) \land (B_1=B)\land (C_1=\paris)  \land \On(B_1,T_1) \land \TIn(T_1,C_1): 8.1; \\
  & \qquad \, \neg: 0]].
  \end{align*}
  Intuitively, this result states that after executing a concrete stochastic $\unload$
  action with arguments $(B_1,T_1,C_1)$, we achieve the highest value (10 plus a discounted 0.9*10) if a box was already in Paris,
  the next highest value (10 occurring with probability 0.9 and discounted by 0.9) if unloading $B_1$ from $T_1$ in $C_1=\paris$, and a
  value of zero otherwise. 
  The main source of efficiency (or lack thereof) of SDP is the ability to perform such operations symbolically and simplify the result into a compact expression.

\item \label{sdp_3} {\bf Object Maximization:} 
Note that up to this point in the algorithm the action arguments are still considered to be concrete arbitrary objects,
$(B_1,T_1,C_1)$ in our example. 
However, we must make sure that in each of the (unspecified and possibly infinite set of possible) states we choose the best concrete action for that state, by specifying the appropriate action arguments. This is handled in the current step of the algorithm.

To achieve this, we maximize over the
  action parameters $X$ of $Q_{V_k}^{A(X)}$ to produce $Q_{V_k}^A$ for each
  action $A(X)$. This implicitly obtains the value achievable by the best
  ground instantiation of $A(X)$ in each state. This step is
  implemented by converting action parameters $X$ 
  to variables, each associated with the $\max$ aggregation operator,
  and appending these operators to the head of the aggregation
  function. Once this is done, further logical simplification may be possible. This occurs in our running example where existential quantification (over $B_1,C_1$) which is constrained by equality can be removed, and the result is:
\begin{align*}
Q_0^{\unload}(S) = & \\
[\max_T, \max_B \, & [\BIn(B,\paris):19;  \\
    & \neg \BIn(B,\paris) \land \On(B,T) \land \TIn(T,\paris): 8.1; \\
    & \neg: 0]].
\end{align*}

\item \label{sdp_4} {\bf Maximize over Actions:} The $k\!+\!1$st step-to-go
  value function $V_{k+1}$ $=$ $\max_A Q_{V_k}^A$, is generated by
  combining the expressions using the binary operation $\max$.
  
  Concretely, for our running example, this means we would compute:
  \begin{align*}
  V_1(S) = \max( Q_0^{\unload}(S), \max( Q_0^{\load}(S), Q_0^{\drive}(S) ) ).
  \end{align*}
  While we have only shown $Q_0^{\unload}(S)$ above, we remark that
  the values achievable in each state by $Q_0^{\unload}(S)$ dominate
  or equal the values achievable by $Q_0^{\load}(S)$ and $Q_0^{\drive}(S)$
  in the same state.  Practically this implies that after simplification
  we obtain the following value function:
  \begin{align*}
    V_1(S) = Q_0^{\unload}&(S) = \\
[\max_T, \max_B & [\BIn(B,\paris):19;  \\
    & \neg \BIn(B,\paris) \land \On(B,T) \land \TIn(T,\paris): 8.1; \\
    & \neg: 0]].    
  \end{align*}
  Critically for the objectives of lifted
  stochastic planning, we observe that the value function derived by
  SDP is indeed lifted: it holds for any number of boxes, trucks and cities.
  
\end{enumerate}

SDP repeats these steps to the required depth, iteratively calculating 
$V_k$.  For example, Figure~\ref{fig:vfun_and_policy} illustrates $V_\infty$
for the \textsc{BoxWorld} example, which was computed by terminating the SDP
loop once the value function converged.

The basic SDP algorithm is an exact calculation whenever the model can
be specified using the constraints above and the reward function can
be specified with $\max$ and $\min$ aggregation \cite{JoshiKeKh11}. 
This is satisfied by
classical models of stochastic planning.  As illustrated, in these cases, the SDP solution conforms to our definition of {generalized lifted inference}.

\subfour{Extending the Scope of SDP}  
The algorithm above cannot handle models with more complex dynamics and rewards as motivated in the introduction. In particular, prior work has considered two important properties that appear to be relevant in many domains. The first is additive rewards, illustrated for example, in Eq~\ref{eq:reward-additive}.
The second property is exogenous branching transitions illustrated above by the disappearing blocks example. 
These represent two different challenges for the SDP algorithm. The first is that we must handle sum aggregation in value functions, despite the fact that this means that some of the operations are not {\em safe} and hence require a special implementation. The second is in modeling the exogenous branching dynamics which requires getting around potential conflicts among such events and between such events and agent actions. 
The introduction illustrated the type of solution that can be expected in such a problem where counting expressions, that measure the number of times certain conditions hold in a state, determine the value in that state. 
 
To date, exact abstract solutions for problems of this form have not been obtained.
The work of \cite{sanner:icaps07}
and 
\cite{Sanner08} (Ch. 6) considered additive rewards and
has formalized an expressive family of models with exogenous events. This work 
has
shown that some specific challenging domains can be handled using several algorithmic ideas, but did not provide a general algorithm that is applicable across problems in this class. 
The work of \cite{JoshiKhRaTaFe13} 
developed a model for ``service domains" which significantly constrains the type of exogenous branching. In their model, a transition includes an agent step whose dynamics use endogenous branching, followed by ``nature's step" where each object (e.g., a box) experiences a random exogenous action (potentially disappearing). 
Given these assumptions, they provide a generally applicable approximation algorithm as follows.
Their algorithm treats agent's actions exactly as in SDP above. To regress nature's actions we follow the following three steps: (1) the summation variables
are first ground using a Skolem constant $c$, then (2) a single exogenous event centered at $c$ is regressed using the same machinery, and finally (3) the Skolemization is reversed to yield another additive value function. 
The complete details are beyond the scope of this chapter.
The algorithm yields a solution that avoids counting formulas and is syntactically close to the one given by the original algorithm. Since such formulas are necessary, the result is an approximation but it was shown to be a conservative one in that it provides a monotonic lower bound on the true value. 
Therefore, this algorithm
conforms to our definition of {\em approximate generalized lifted inference}. 

In our example, starting with the reward of Eq~(\ref{eq:reward-additive}) we first replace the sum aggregation with a scaled version of average aggregation (which is safe w.r.t.\ summation)
\begin{equation*}
[n \cdot \mbox{avg}_B [\BIn(B,\paris): 10; \neg: 0]]
\end{equation*}
and then ground it to get
\begin{equation*}
[n \cdot [\BIn(c,\paris): 10; \neg: 0]].
\end{equation*}
The next step is to regress through the exogenous event at $c$. The problem where boxes disappear with probability 0.2 can be cast as having two action variants where ``disappearing-block" succeeds with probability 0.2 and fails with probability 0.8.
Regressing the success variant we get the expression $[0]$ (the zero function) and regressing the fail variant we get
$[n \cdot [\BIn(c,\paris): 10; \neg: 0]]$. Multiplying by the probabilities of the variants we get:
$[0]$ and  $[n \cdot [\BIn(c,\paris): 8; \neg: 0]]$ and adding them (there are no variables to standardize apart) we get
\begin{equation*}
[n \cdot  [\BIn(c,\paris): 8; \neg: 0]].
\end{equation*}
Finally lifting the last equation we get
\begin{equation*}
[n \cdot \mbox{avg}_B [\BIn(B,\paris): 8; \neg  \BIn(B,\paris): 0]].
\end{equation*}
Next we follow with the standard steps of SDP for the agent's action. The steps are analogous to the example of SDP given above. 
Considering the discussion in the 
introduction (recall that in order to simplify the reasoning in this case we omitted discounting and adding the reward) this algorithm produces 
  \begin{align*}
& [n \cdot  \max_T, \mbox{avg}_B, 
[\BIn(B,\paris):8;  \\
& (\neg \BIn(B,\paris) \land \On(B,T) \land \TIn(T,\paris)): 7.2; \neg: 0]] , 
  \end{align*}
which is identical to the exact expression given in the introduction.
As already mentioned, the result is not guaranteed to be exact in general. 
In addition, the maximization in step~iv of SDP requires some ad-hoc implementation because maximization is not safe with respect to average aggregation.

It is clear from the above example that the main difficulty in extending SDP is due to the
interaction of the counting formulas arising from exogenous events and
additive rewards with the first-order aggregation structure inherent
in the planning problem.  Relational expressions, their GFODD counterparts, and other representations that have been used to date are not able to combine these effectively. A representation that seamlessly supports both
relational expressions and operations on them along with counting expressions
might allow for more robust versions of generalized lifted inference to be realized.

\section{Discussion and Related Work}

As motivated in the introduction, SDP has explored probabilistic inference problems with a specific form of alternating maximization and expectation blocks. The main computational advantage comes from lifting in the sense of lifted inference in standard first order logic. Issues that arise from conditional summations over combinations random variables,  common in probabilistic lifted inference, have been touched upon but not extensively. In cases where SDP has been shown to work it provides {\em generalized lifted inference} where the complexity of the inference algorithm is completely independent of the domain size (number of objects) in problem specification, and where the response to queries is either independent of that size or can be specified parametrically. 
This is a desirable property but to our knowledge it is not shared by most work on probabilistic lifted inference. A notable exception is given by the knowledge compilation result of \cite{vandenbroeck-thesis} (see Chapter 4 and Theorem 5.5) 
and the recent work in \cite{KazemiP16,KazemiKBP16}, where a model is compiled into an alternative form parametrized by the domain $D$ and where responses to queries can be obtained in polynomial time as a function of $D$. 
The emphasis in that work is on being {\em domain lifted} (i.e., being polynomial in domain size). Generalized lifted inference requires an algorithm whose results can be computed once, in time independent of that size, and then reused to evaluate the answer for specific domain sizes.
This analogy also shows that SDP can be seen as a compilation algorithm, compiling a domain model into  a more accessible form representing the value function, which can be queried efficiently. 
This connection provides an interesting new perspective on both fields.

In this chapter we focused on one particular instance of SDP. 
Over the last 15 years SDP has seen a significant amount of work expanding over the original algorithm 
by using different representations, by using algorithms other than value iteration, and by extending the models and algorithms to more complex settings. In addition, several ``lifted" inductive approaches that do not strictly fall within the probabilistic inference paradigm have been developed. 
We review this work in the remainder of this section.

\subsection{Deductive Lifted Stochastic Planning}

As a precursor to its use in lifted stochastic planning, the term SDP
originated in the propositional logical
context~\cite{bout-dean-hanks,boutilier99dt} when it was realized that
propositionally structured MDP transitions (i.e., dynamic Bayesian
networks~\cite{dbn}) and rewards (e.g., trees that exploited
context-specific independence~\cite{csi}) could be used to define
highly compact \textit{factored MDPs}; this work also realized that the
factored MDP structure could be exploited for representational
compactness and computational efficiency by leveraging symbolic
representations (e.g., trees) in dynamic programming.  Two highly
cited (and still used algorithms) in this area of work are the
SPUDD~\cite{spudd} and APRICODD~\cite{apricodd} algorithms that
leveraged algebraic decision diagrams (ADDs)~\cite{BaharFrGaHaMaPaSo93} for,
respectively, exact and approximate solutions to factored MDPs.
Recent work in this area \cite{lesner:ppddl11} shows how to perform propositional SDP 
directly with  ground representations in PPDDL~\cite{ppddl}, and develops extensions
for factored action spaces \cite{raghavan2012planning,raghavan2013symbolic}.

Following the seminal introduction of {\it lifted} SDP
in~\cite{BoutilierRePr01}, several early papers on SDP approached the
problem with existential rewards with different representation
languages that enabled efficient implementations. This includes the
First-order value iteration
(FOVIA)~\citep{lao_fovia,HolldoblerKaSk2006}, the Relational Bellman
algorithm (ReBel)~\citep{KerstingOtDe04}, and the FODD based
formulation of \citep{WangJoKh08,JoshiKh08,JoshiKeKh10}.

Along this dimension two representations are closely related to the
relational expression of this chapter.  As mentioned above, relational
expressions are an abstraction of the GFODD representation
\citep{JoshiKeKh11,JoshiKhRaTaFe13,HescottKh15} which captures
expressions using a decision diagram formulation extending
propositional ADDs \cite{BaharFrGaHaMaPaSo93}.  In particular, paths
in the graphical representation of the DAG representing the GFODD
correspond to the mutually exclusive conditions in expressions. The
aggregation in GFODDs and relational expressions provides significant
expressive power in modeling relational MDPs. The GFODD representation
is more compact than relational expressions but requires more complex
algorithms for its manipulation.  The other closely related
representation is the case notation of
\cite{BoutilierRePr01,SannerBo09}.  The case notation is similar to
relational expressions in that we have a set of conditions (these are
mostly in a form that is mutually exclusive but not always so) but the
main difference is that quantification is done within each case
separately, and the notion of aggregation is not fully developed.
First-order algebraic decision diagrams
(FOADDs)~\citep{Sanner08,SannerBo09} are related to the case
notation in that they require closed formulas within diagram nodes,
i.e., the quantifiers are included within the graphical representation
of the expression.  The use of quantifiers inside cases and nodes
allows for an easy incorporation of off the shelf theorem provers for
simplification.
Both FOADD and GFODD were used to extend SDP to capture additive rewards and exogenous events as already discussed in the previous section.
While the representations (relational expression and GFODDs vs.\ case notation and FOADD) have similar expressive power, the difference in aggregation makes for different algorithmic properties that are hard to compare in general. 
However, the modular treatment of aggregation in GFODDs and the generic form of operations over them makes them the most flexible alternative to date for directly manipulating the aggregated case representation used in this chapter.

The idea of SDP has also been extended in terms of the choice of
planning algorithm, as well as to the case of partially observable
MDPs.  Case notation and FOADDs have been used to implement
approximate linear programming~\citep{foalp,SannerBo09} and
approximate policy iteration via linear programming~\citep{foapi} and
FODDs have been used to implement relational policy iteration
\cite{WangKh07}.  GFODDs have also been used for open world reasoning
and applied in a robotic context \cite{JoshiSKS12}.  The work of
\cite{WangK10} and \cite{SannerK10} explore SDP solutions, with GFODDs
and case notation respectively, to relational partially observable MDPs (POMDPs) where the
problem is conceptually and algorithmically much more complex.
Related work in POMDPs has not explicitly addressed SDP, but rather has
implicitly addressed lifted solutions through the identification of (and
abstraction over) symmetries in applications of dynamic programming
for POMDPs~\cite{doshi:permpomdp08,kim:sympomdp12}.

\subsection{Inductive Lifted Stochastic Planning}

Inductive methods can be seen to be orthogonal to the inference algorithms in that they mostly do not require a model and do not reason about that model. However, 
the overall objective of
producing lifted value functions and policies is shared with the
previously discussed deductive approaches.  
We therefore review these here for completeness. As we discuss, it is also possible  
to combine the inductive and deductive approaches in several ways.

The basic inductive approaches learn a policy directly from a teacher, sometimes known as behavioral cloning. 
The work of \cite{Khardon96,Khardon99,givan:uai02} provided learning algorithms for relational policies with theoretical and empirical evidence for their success. 
Relational policies and value functions were also explored in reinforcement learning.
This
was done with pure reinforcement learning using relational 
regression trees to learn a 
Q-function~\citep{dzeroski01},
combining this with supervised guidance~\citep{driessens02}, or using
Gaussian processes and graph kernels over relational structures to
learn a 
Q-function~\citep{driessens06}.
A more recent approach uses 
functional gradient boosting with lifted
regression trees to learn lifted policy structure in
a policy gradient algorithm~\cite{kersting:nppg}.

Finally, several approaches combine inductive and deductive elements. 
The work of 
\cite{gretton_thiebaux} combines inductive logic programming with first-order
decision-theoretic regression, by first using deductive methods (decision theoretic regression) to
generate candidate policy structure, and then learning using this structure as features. 
The work of  \citep{givan:jair06} shows how one can implement relational approximate policy iteration where policy improvement steps are performed by learning the intended policy from generated trajectories instead of direct calculation. 
Although these approaches are partially deductive they do not share the common theme of this chapter relating planning and inference in relational contexts.

\section{Conclusions}

This chapter provides a review of SDP methods, that perform abstract reasoning for stochastic planning, from the viewpoint of probabilistic inference. We have illustrated how the planning problem and the inference problem are related. Specifically, finite horizon optimization in MDPs is related to an inference problem with alternating maximization and expectation blocks and is therefore more complex than  marginal MAP queries that have been studied in the literature. This analogy is valid both at the propositional and relational levels and it suggests a new line of challenges for inference problems in discrete domains.We have also identified the opportunity for generalized lifted inference, where the algorithm and its solution are  agnostic of the domain instance and its size and are efficient regardless of this size. We have shown that under some conditions SDP algorithms provide generalized lifted inference. In more complex models, especially ones with additive rewards and exogenous events, SDP algorithms are yet to mature into an effective and widely applicable inference scheme. On the other hand, the challenges faced in such problems are exactly the ones typically seen in standard lifted inference problems. Therefore, exploring generalized lifted inference more abstractly has the potential to lead to advances in both areas.

\section*{Acknowledgments} 
This work is partly supported by NSF grants IIS-0964457 and IIS-1616280.

\bibliographystyle{unsrt}
\bibliography{foddbib,sanner}

\begin{thebibliography}{10}

\bibitem{BoutilierRePr01}
C.~Boutilier, R.~Reiter, and B.~Price.
\newblock Symbolic dynamic programming for first-order {MDPs}.
\newblock In {\em Proc. of IJCAI}, pages 690--700, 2001.

\bibitem{strips}
Richard~E. Fikes and Nils~J. Nilsson.
\newblock {STRIPS}: A new approach to the application of theorem proving to
  problem solving.
\newblock {\em AI Journal}, 2:189--208, 1971.

\bibitem{pso}
Neil Kushmerick, Steve Hanks, and Dan Weld.
\newblock An algorithm for probabilistic planning.
\newblock {\em Artificial Intelligence}, 76(12):239--286, 1995.

\bibitem{McCarthy58}
J.~McCarthy.
\newblock Programs with common sense.
\newblock In {\em Proceedings of the Symposium on the Mechanization of Thought
  Processes}, volume~1, pages 77--84. National Physical Laboratory, 1958.
\newblock Reprinted in R. Brachman and H. Levesque (Eds.), Readings in
  Knowledge Representation, 1985, Morgan Kaufmann, Los Altos, CA.

\bibitem{Attias03}
Hagai Attias.
\newblock Planning by probabilistic inference.
\newblock In {\em Proceedings of the Ninth International Workshop on Artificial
  Intelligence and Statistics, {AISTATS} 2003, Key West, Florida, USA, January
  3-6, 2003}, 2003.

\bibitem{ToussaintSt06}
M.~Toussaint and A.~Storsky.
\newblock Probabilistic inference for solving discrete and continuous sta te
  markov decision processes.
\newblock In {\em Proceedings of the International Conference on Machine
  Learning}, 2006.

\bibitem{domshlak2006}
Carmel Domshlak and J{\"o}rg Hoffmann.
\newblock Fast probabilistic planning through weighted model counting.
\newblock In {\em Proceedings of the International Conference on Automated
  Planning and Scheduling}, 2006.

\bibitem{LangTo09}
M.~Lang and M.~Toussaint.
\newblock Approximate inference for planning in stochastic relational worlds.
\newblock In {\em Proceedings of the International Conference on Machine
  Learning}, 2009.

\bibitem{FurmstonB10}
Thomas Furmston and David Barber.
\newblock Variational methods for reinforcement learning.
\newblock In {\em Proceedings of the International Conference on Artificial
  Intelligence and Statistics, {AISTATS}}, pages 241--248, 2010.

\bibitem{LiuI12}
Qiang Liu and Alexander~T. Ihler.
\newblock Belief propagation for structured decision making.
\newblock In {\em Proceedings of the Conference on Uncertainty in Artificial
  Intelligence (UAI)}, pages 523--532, 2012.

\bibitem{ChengLCI13}
Qiang Cheng, Qiang Liu, Feng Chen, and Alexander~T. Ihler.
\newblock Variational planning for graph-based mdps.
\newblock In {\em International Conference on Neural Information Processing
  Systems}, pages 2976--2984, 2013.

\bibitem{LeeMaDe14}
J.~Lee, R.~Marinescau, and R.~Dechter.
\newblock Applying marginal map search to probabilistic conformant planning.
\newblock In {\em Fourth International Workshop on Statistical Relational AI
  (StarAI)}, 2014.

\bibitem{LeeMaDe16}
J.~Lee, R.~Marinescau, and R.~Dechter.
\newblock Applying search based probabilistic inference algorithms to
  probabilistic conformant planning: Preliminary results.
\newblock In {\em Proceedings of the International Symposium on Artificial
  Intelligence and Mathematics (ISAIM)}, 2016.

\bibitem{issakkimuthu2015hop}
M.~Issakkimuthu, A.~Fern, R.~Khardon, P.~Tadepalli, and S.~Xue.
\newblock Hindsight optimization for probabilistic planning with factored
  actions.
\newblock In {\em ICAPS}, 2015.

\bibitem{MeentPTW16}
Jan{-}Willem van~de Meent, Brooks Paige, David Tolpin, and Frank Wood.
\newblock Black-box policy search with probabilistic programs.
\newblock In {\em Proceedings of the International Conference on Artificial
  Intelligence and Statistics, {AISTATS}}, pages 1195--1204, 2016.

\bibitem{JoshiKeKh11}
S.~Joshi, K.~Kersting, and R.~Khardon.
\newblock Decision theoretic planning with generalized first order decision
  diagrams.
\newblock {\em AIJ}, 175:2198--2222, 2011.

\bibitem{JoshiKhRaTaFe13}
S.~Joshi, R.~Khardon, A.~Raghavan, P.~Tadepalli, and A.~Fern.
\newblock Solving relational {MDPs} with exogenous events and additive rewards.
\newblock In {\em ECML}, 2013.

\bibitem{SannerBo09}
S.~Sanner and C.~Boutilier.
\newblock Practical solution techniques for first order {MDP}s.
\newblock {\em AIJ}, 173:748--788, 2009.

\bibitem{Lloyd87}
J.W. Lloyd.
\newblock {\em Foundations of Logic Programming}.
\newblock Springer Verlag, 1987.
\newblock Second Edition.

\bibitem{RussellNo95}
S.~Russell and P.~Norvig.
\newblock {\em Artificial Intelligence: a modern approach}.
\newblock Prentice Hall, 1995.

\bibitem{ChangKe90}
C.~Chang and J.~Keisler.
\newblock {\em Model Theory}.
\newblock Elsevier, Amsterdam, Holland, 1990.

\bibitem{RichardsonDo06}
M.~Richardson and P.~Domingos.
\newblock Markov logic networks.
\newblock {\em Machine Learning}, 62:107--136, 2006.

\bibitem{Problog}
L.~De Raedt, A.~Kimmig, and H.~Toivonen.
\newblock Problog: A probabilistic prolog and its application in link
  discovery.
\newblock In {\em Proc. of the IJCAI}, pages 2462--2467, 2007.

\bibitem{LiftedWMC}
G.~Van den Broeck, N.~Taghipour, W.~Meert, J.~Davis, and L.~De Raedt.
\newblock Lifted probabilistic inference by first-order knowledge compilation.
\newblock In {\em Proc. of the IJCAI}, pages 2178--2185, 2011.

\bibitem{RussellNo09}
S.~Russell and P.~Norvig.
\newblock {\em Artificial Intelligence: a modern approach}.
\newblock Prentice Hall, 2009.
\newblock 3rd Edition.

\bibitem{Puterman1994}
M.~L. Puterman.
\newblock {\em Markov decision processes: Discrete stochastic dynamic
  programming}.
\newblock Wiley, 1994.

\bibitem{WangJoKh08}
C.~Wang, S.~Joshi, and R.~Khardon.
\newblock First order decision diagrams for relational {MDP}s.
\newblock {\em JAIR}, 31:431--472, 2008.

\bibitem{reiter_KIA}
Ray Reiter.
\newblock {\em Knowledge in Action: Logical Foundations for Specifying and
  Implementing Dynamical Systems}.
\newblock MIT Press, 2001.

\bibitem{Sanner08}
S.~Sanner.
\newblock {\em First-order decision-theoretic planning in structured relational
  environments}.
\newblock PhD thesis, University of Toronto, 2008.

\bibitem{sanner:icaps07}
S.~Sanner and C.~Boutilier.
\newblock Approximate solution techniques for factored first-order {MDP}s.
\newblock In {\em Proceedings of the 17th Conference on Automated Planning and
  Scheduling ({ICAPS-07})}, 2007.

\bibitem{vandenbroeck-thesis}
Guy Van~den Broeck.
\newblock {\em Lifted Inference and Learning in Statistical Relational Models}.
\newblock PhD thesis, KU Leuven, 2013.

\bibitem{KazemiP16}
Seyed~Mehran Kazemi and David Poole.
\newblock Knowledge compilation for lifted probabilistic inference: Compiling
  to a low-level language.
\newblock In {\em Proceedings of the Conference on Knowledge Representation and
  Reasoning}, pages 561--564, 2016.

\bibitem{KazemiKBP16}
Seyed~Mehran Kazemi, Angelika Kimmig, Guy~Van den Broeck, and David Poole.
\newblock New liftable classes for first-order probabilistic inference.
\newblock In {\em International Conference on Neural Information Processing
  Systems}, pages 3117--3125, 2016.

\bibitem{bout-dean-hanks}
Craig Boutilier, Thomas Dean, and Steve Hanks.
\newblock Planning under uncertainty: Structural assumptions and computational
  leverage.
\newblock In {\em Third European Workshop on Planning}, Assisi, Italy, 1995.

\bibitem{boutilier99dt}
Craig Boutilier, Thomas Dean, and Steve Hanks.
\newblock Decision-theoretic planning: Structural assumptions and computational
  leverage.
\newblock {\em Journal of Artificial Intelligence Research ({JAIR})}, 11:1--94,
  1999.

\bibitem{dbn}
Thomas Dean and Keiji Kanazawa.
\newblock A model for reasoning about persistence and causation.
\newblock {\em Computational Intelligence}, 5(3):142--150, 1989.

\bibitem{csi}
Craig Boutilier, Nir Friedman, Mois\'{e}s Goldszmidt, and Daphne Koller.
\newblock Context-specific independence in {Bayesian} networks.
\newblock In {\em Uncertainty in Artificial Intelligence ({UAI-96})}, pages
  115--123, Portland, OR, 1996.

\bibitem{spudd}
Jesse Hoey, Robert St-Aubin, Alan Hu, and Craig Boutilier.
\newblock {SPUDD}: Stochastic planning using decision diagrams.
\newblock In {\em Uncertainty in Artificial Intelligence ({UAI-99})}, pages
  279--288, Stockholm, 1999.

\bibitem{apricodd}
Robert St-Aubin, Jesse Hoey, and Craig Boutilier.
\newblock {APRICODD}: Approximate policy construction using decision diagrams.
\newblock In {\em Advances in Neural Information Processing 13 ({NIPS-00})},
  pages 1089--1095, Denver, 2000.

\bibitem{BaharFrGaHaMaPaSo93}
R.~Bahar, E.~Frohm, C.~Gaona, G.~Hachtel, E.~Macii, A.~Pardo, and F.~Somenzi.
\newblock Algebraic decision diagrams and their applications.
\newblock In {\em IEEE /ACM ICCAD}, pages 188--191, 1993.

\bibitem{lesner:ppddl11}
Boris Lesner and Bruno Zanuttini.
\newblock Efficient policy construction for mdps represented in probabilistic
  pddl.
\newblock In Fahiem Bacchus, Carmel Domshlak, Stefan Edelkamp, and Malte
  Helmert, editors, {\em ICAPS}. AAAI, 2011.

\bibitem{ppddl}
H{\aa}kan L.~S. Younes, Michael~L. Littman, David Weissman, and John Asmuth.
\newblock The first probabilistic track of the international planning
  competition.
\newblock {\em Journal of Artificial Intelligence Research ({JAIR})},
  24:851--887, 2005.

\bibitem{raghavan2012planning}
Aswin Raghavan, Saket Joshi, Alan Fern, Prasad Tadepalli, and Roni Khardon.
\newblock Planning in {F}actored {A}ction {S}paces with {S}ymbolic {D}ynamic
  {P}r ogramming.
\newblock In {\em Proceedings of the AAAI Conference on Artificial
  Intelligence}, 2012.

\bibitem{raghavan2013symbolic}
Aswin Raghavan, Roni Khardon, Alan Fern, and Prasad Tadepalli.
\newblock Symbolic opportunistic policy iteration for factored-action mdps.
\newblock In {\em International Conference on Neural Information Processing
  Systems}, pages 2499--2507, 2013.

\bibitem{lao_fovia}
Eldar Karabaev and Olga Skvortsova.
\newblock A heuristic search algorithm for solving first-order {MDP}s.
\newblock In {\em Uncertainty in Artificial Intelligence ({UAI-05})}, pages
  292--299, Edinburgh, Scotland, 2005.

\bibitem{HolldoblerKaSk2006}
S.~H{\"o}lldobler, E.~Karabaev, and O.~Skvortsova.
\newblock {FluCaP:} a heuristic search planner for first-order {MDPs}.
\newblock {\em JAIR}, 27:419--439, 2006.

\bibitem{KerstingOtDe04}
K.~Kersting, M.~Van Otterlo, and L.~{De Raedt}.
\newblock Bellman goes relational.
\newblock In {\em Proc. of ICML}, 2004.

\bibitem{JoshiKh08}
S.~Joshi and R.~Khardon.
\newblock Stochastic planning with first order decision diagrams.
\newblock In {\em Proc. of ICAPS}, pages 156--163, 2008.

\bibitem{JoshiKeKh10}
S.~Joshi, K.~Kersting, and R.~Khardon.
\newblock Self-taught decision theoretic planning with first order decision
  diagrams.
\newblock In {\em Proc. of ICAPS}, pages 89--96, 2010.

\bibitem{HescottKh15}
B.~Hescott and R.~Khardon.
\newblock The complexity of reasoning with {FODD} and {GFODD}.
\newblock {\em Artificial Intelligence}, 2015.

\bibitem{foalp}
Scott Sanner and Craig Boutilier.
\newblock Approximate linear programming for first-order {MDP}s.
\newblock In {\em Uncertainty in Artificial Intelligence ({UAI-05})}, pages
  509--517, Edinburgh, Scotland, 2005.

\bibitem{foapi}
Scott Sanner and Craig Boutilier.
\newblock Practical linear evaluation techniques for first-order {MDP}s.
\newblock In {\em Uncertainty in Artificial Intelligence ({UAI-06})}, Boston,
  Mass., 2006.

\bibitem{WangKh07}
C.~Wang and R.~Khardon.
\newblock Policy iteration for relational {MDPs}.
\newblock In {\em Proceedings of UAI}, 2007.

\bibitem{JoshiSKS12}
Saket Joshi, Paul~W. Schermerhorn, Roni Khardon, and Matthias Scheutz.
\newblock Abstract planning for reactive robots.
\newblock In {\em ICRA}, pages 4379--4384, 2012.

\bibitem{WangK10}
Chenggang Wang and Roni Khardon.
\newblock Relational partially observable {MDPs}.
\newblock In {\em Proceedings of the Twenty-Fourth {AAAI} Conference on
  Artificial Intelligence, {AAAI} 2010, Atlanta, Georgia, USA, July 11-15,
  2010}, 2010.

\bibitem{SannerK10}
Scott Sanner and Kristian Kersting.
\newblock Symbolic dynamic programming for first-order {POMDPs}.
\newblock In {\em Proceedings of the Twenty-Fourth {AAAI} Conference on
  Artificial Intelligence, {AAAI} 2010, Atlanta, Georgia, USA, July 11-15,
  2010}, 2010.

\bibitem{doshi:permpomdp08}
Finale Doshi and Nicholas Roy.
\newblock The permutable pomdp: Fast solutions to pomdps for preference
  elicitation.
\newblock In {\em Proceedings of the Seventh International Conference on
  Autonomous Agents and Multiagent Systems (AAMAS 2008)}, Estoril, Portugal,
  May 2008.

\bibitem{kim:sympomdp12}
Byung~Kon Kang and Kee{-}Eung Kim.
\newblock Exploiting symmetries for single- and multi-agent partially
  observable stochastic domains.
\newblock {\em Artif. Intell.}, 182-183:32--57, 2012.

\bibitem{Khardon96}
R.~Khardon.
\newblock Learning to take actions.
\newblock {\em Machine Learning}, 35:57--90, 1999.

\bibitem{Khardon99}
R.~Khardon.
\newblock Learning action strategies for planning domains.
\newblock {\em Artificial Intelligence}, 113(1-2):125--148, 1999.

\bibitem{givan:uai02}
SungWook Yoon, Alan Fern, and Robert Givan.
\newblock Inductive policy selection for first-order {Markov} decision
  processes.
\newblock In {\em Uncertainty in Artificial Intelligence ({UAI-02})}, pages
  569--576, Edmonton, 2002.

\bibitem{dzeroski01}
Saso Dzeroski, Luc {DeRaedt}, and Kurt Driessens.
\newblock Relational reinforcement learning.
\newblock {\em Machine Learning Journal ({MLJ})}, 43:7--52, 2001.

\bibitem{driessens02}
Kurt Driessens and Saso Dzeroski.
\newblock Integrating experimentation and guidance in relational reinforcement
  learning.
\newblock In {\em International Conference on Machine Learning ({ICML})}, pages
  115--122, 2002.

\bibitem{driessens06}
Thomas Gartner, Kurt Driessens, and Jan Ramon.
\newblock Graph kernels and gaussian processes for relational reinforcement
  learning.
\newblock {\em Machine Learning Journal ({MLJ})}, 64:91--119, 2006.

\bibitem{kersting:nppg}
Kristian Kersting and Kurt Driessens.
\newblock Non-parametric policy gradients: A unified treatment of propositional
  and relational domains.
\newblock In {\em Proceedings of the 25th International Conference on Machine
  Learning}, ICML '08, pages 456--463, New York, NY, USA, 2008. ACM.

\bibitem{gretton_thiebaux}
Charles Gretton and Sylvie Thiebaux.
\newblock Exploiting first-order regression in inductive policy selection.
\newblock In {\em Uncertainty in Artificial Intelligence ({UAI-04})}, pages
  217--225, Banff, Canada, 2004.

\bibitem{givan:jair06}
Sungwook Yoon, Alan Fern, and Robert Givan.
\newblock Approximate policy iteration with a policy language bias: Learning to
  solve relational markov decision processes.
\newblock {\em Journal of Artificial Intelligence Research ({JAIR})},
  25:85--118, 2006.

\end{thebibliography}

\end{document}